\documentclass{article}

\usepackage[final]{neurips_2025}

\usepackage[utf8]{inputenc} 
\usepackage[T1]{fontenc}    
\usepackage{hyperref}       
\usepackage{url}            
\usepackage{booktabs}       
\usepackage{amsfonts}       
\usepackage{nicefrac}       
\usepackage{microtype}      
\usepackage{xcolor}         
\usepackage{graphicx}
\usepackage{multirow}
\usepackage{siunitx}
\usepackage{amsmath}
\usepackage{wasysym}
\usepackage{subcaption}
\usepackage{float}
\usepackage{makecell}
\usepackage{amssymb}
\usepackage{dsfont}
\usepackage{bm}
\usepackage{array}
\usepackage{wrapfig}
\newcommand{\tableCellHeight}{1.1}
\newcommand{\tabstyle}[1]{
  \setlength{\tabcolsep}{#1}
  \renewcommand{\arraystretch}{\tableCellHeight}
  \centering
  \footnotesize
}

\title{Mitigating Hallucination in Multimodal LLMs with Layer Contrastive Decoding}

\author{
    Bingkui Tong\textsuperscript{1}\quad 
    Jiaer Xia\textsuperscript{2}\quad
    Kaiyang Zhou\textsuperscript{2}\\
    \textsuperscript{1}Mohamed bin Zayed University of Artificial Intelligence\\
    \textsuperscript{2}Hong Kong Baptist University
}

\begin{document}

\maketitle

\begin{abstract}
Multimodal Large Language Models (MLLMs) have shown impressive perception and reasoning capabilities, yet they often suffer from hallucinations—generating outputs that are linguistically coherent but inconsistent with the context of the input image, including inaccuracies in objects, attributes, and relations.
To address this challenge, we propose a simple approach called Layer Contrastive Decoding (LayerCD). Our design is motivated by the observation that shallow visual features are much more likely than deep visual features to cause an MLLM to hallucinate as they only capture biased, low-level information that is insufficient for high-level reasoning. Therefore, LayerCD aims to filter out hallucinations by contrasting the output distributions generated from visual features of different levels, specifically those from the shallow and deep layers of the vision encoder, respectively.
%
%
We conduct extensive experiments on two hallucination benchmarks and show that LayerCD significantly outperforms current state-of-the-art. The code for LayerCD is available at \href{https://github.com/maifoundations/LayerCD}{maifoundations/LayerCD}.
\end{abstract}    
\section{Introduction}
\label{sec:introduction}




\begin{wrapfigure}{r}{0.5\textwidth} 
    \vspace{-10pt} 
    \centering
    \includegraphics[width=\linewidth]{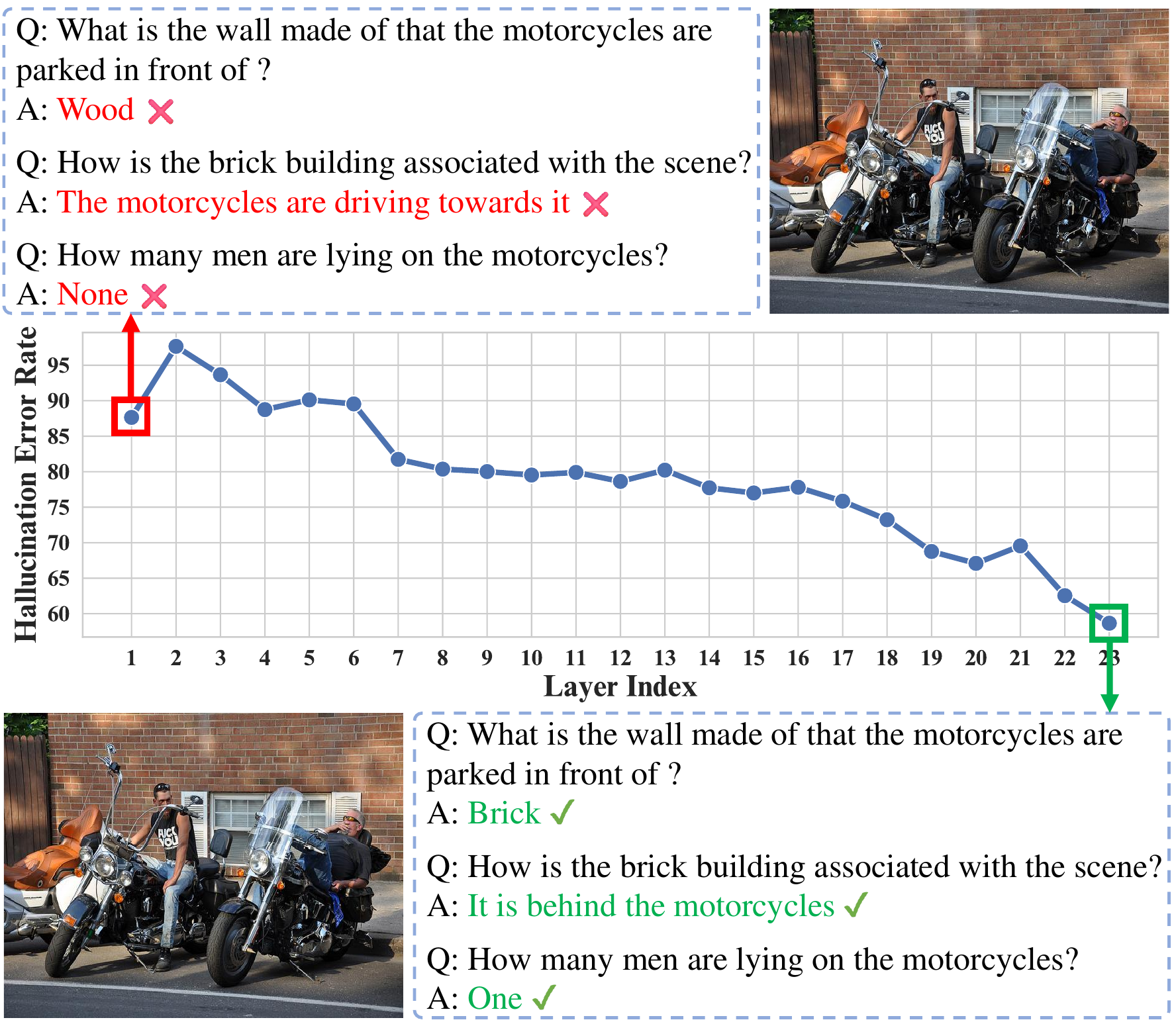} 
    \caption{Evaluation of LLaVA 1.5, a state-of-the-art MLLM, using different levels of visual features. The results suggest that shallow features lead to significantly higher hallucination error rates compared to deeper features.}
    \label{fig:Intuition}
    \vspace{-10pt} 
\end{wrapfigure}

LLMs have long struggled with hallucination, a phenomenon where model outputs appear plausible but are factually incorrect or fabricated~\cite{maynez2020faithfulness}. Unfortunately, multimodal LLMs (MLLMs)---which incorporate an additional vision module to process images---also face this issue. In MLLMs, hallucination occurs when the model generates responses that are fluent and coherent yet misaligned with the visual input~\cite{liu2024asurvey}. These inconsistencies often manifest as inaccuracies in identifying objects, attributes, or relationships, limiting the model's ability to accurately interpret images and posing a significant challenge for real-world deployment and practical applications.

\begin{figure*}[!t]
    \centering
    \includegraphics[width=0.9\textwidth]{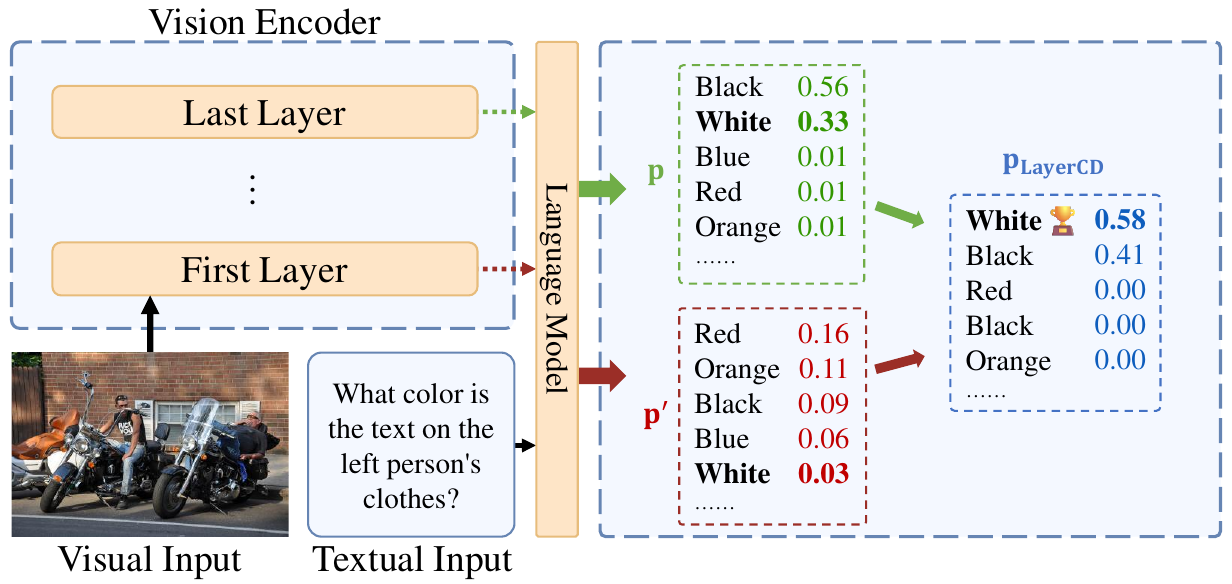}
    \caption{Overview of Layer Contrastive Decoding (LayerCD). The main idea is to factor out hallucinations by contrasting output distributions derived from different levels of visual features, specifically those from the shallow and deep layers of the vision encoder, respectively.}
    \label{fig:overview}
\end{figure*}

We propose Layer Contrastive Decoding (LayerCD), a simple, inference-time method that requires no architectural changes. Our approach is motivated by the key observation that MLLMs are more prone to hallucination when conditioned on shallow versus deep visual features. By contrasting the output distributions from these two feature levels, LayerCD effectively reduces hallucinations. A preliminary study using LLaVA 1.5~\cite{liu2024improved} validates this; as shown in Fig.~\ref{fig:Intuition}, using shallow features leads to significantly higher hallucination error rates. This is because shallow features capture only low-level characteristics like edges and colors, which are insufficient for high-level reasoning and thus more likely to cause hallucinations~\cite{zeiler2014visualizing,yosinski2014transferable}.

Fig.~\ref{fig:overview} illustrates the LayerCD mechanism and the crucial differences in the resulting output distributions. While a model using deep features may produce hallucinations (e.g., ``Black'', Blue''), it still assigns non-trivial probability to the correct token (e.g., ``White''). In contrast, conditioning on shallow features yields more high-confidence hallucinations while suppressing the correct token. Drawing inspiration from Contrastive Decoding~\cite{li2022contrastive}, which was designed to reduce repetition in text, LayerCD contrasts these two distributions. This readjustment cancels out hallucinations and allows the correct token (``White'') to emerge with the highest confidence.

In summary, we make the following contributions in this paper: (1) We provide an important insight that shallow visual features are more prone to causing hallucinations in MLLMs than deep visual features; (2) Based on the insight, we propose a simple contrastive decoding approach that contrasts visual features of different layers to factor out hallucinations; (3) We demonstrate the effectiveness of our approach on two hallucination benchmarks where our approach significantly outperforms current state-of-the-art.

\section{Methodology}
\label{sec:method}

Our proposed Layer Contrastive Decoding (LayerCD) is an inference-time, architecture-free method to mitigate hallucination. It is motivated by the observation that contrasting the outputs from shallow visual features (high confidence on hallucinations, low on correct tokens) and deep features (high confidence on both) can isolate and suppress erroneous tokens. During decoding, an MLLM takes visual features $\bm{z}$ and text input $\bm{x}$ to produce the next-token probability, conditioned on previous tokens $\bm{y}_{<t}$:
$
p(y_t | \bm{x}, \bm{z},\bm{y}_{<t})
$

\subsection{Layer Contrastive Decoding}
\paragraph{Contrastive Probability Distribution}
As discussed previously, shallow visual features are more prone than deep features to cause hallucinations and therefore the two output distributions derived from these two types of features can form a contrastive pair, which allows the model to pinpoint and then cancel out hallucinations in the output. Let $\bm{z}_{s}$ denote image features extracted from a shallow layer (e.g., the first layer) in the vision encoder, $\bm{z}_{d}$ image features extracted from a deep layer (e.g., the last layer), and $f(\cdot)$ the LLM, the contrastive probability distribution can be formulated as
\begin{equation} \label{eq:contrastive_prob}
p(\bm{y}|\bm{x}, \bm{z}_d, \bm{z}_s) = \sigma [ (1+\alpha) f(\bm{x}, \bm{z}_d) - \alpha f(\bm{x}, \bm{z}_s) ],
\end{equation}
where $\sigma[\cdot]$ is the softmax function, and $\alpha$ is a hyperparameter that controls the amplification of the difference between the two distributions. In particular, a larger $ \alpha $ indicates a stronger contrast between the two distributions. When $\alpha \! = \! 0$, the formulation returns to regular decoding.

Since LayerCD only modifies the next-token probability distribution, it can be easily combined with existing decoding methods, such as nucleus sampling, beam search, and others. It is worth noting that the formulation in Eq.~\ref{eq:contrastive_prob} does not require modifying the internal model parameters. Unless otherwise specified, we select the first and last vision encoder layer for extracting $\bm{z}_s$ and $\bm{z}_d$, respectively.

\paragraph{Adaptive Plausibility Constraint}
The formulation in Eq.~\ref{eq:contrastive_prob} indiscriminately penalizes all outputs from the shallow-feature model, an assumption that is too strong in practice. Common tokens like articles (e.g., ``a'', ``the'') are often predicted with high confidence regardless of feature depth, and should not be penalized. To prevent the generation of implausible outputs, we introduce an adaptive plausibility constraint, inspired by Li et al.~\cite{li2022contrastive}. At each inference step, we dynamically filter the vocabulary by removing tokens whose confidence in the original deep-feature distribution falls below a threshold based on the maximum confidence. This ensures that LayerCD is computed using only this updated set of plausible tokens:

\begin{equation} \label{eq:constraint}
\begin{aligned}
\mathcal{V}_{\text{head}}(\bm{y}_{<t}) = \{ y_t \in \mathcal{V} : \  p(y_t | \bm{x}, \bm{z}_d, \bm{y}_{<t}) \geq 
\beta\max_{w} \  p(w | \bm{x}, \bm{z}_d, \bm{y}_{<t}) \},
\end{aligned}
\end{equation}
where $ \mathcal{V} $ is the vocabulary of the MLLM and $ \beta $ is a hyperparameter between 0 and 1 that controls how aggressively low-confidence tokens are pruned. A larger $ \beta $ results in a more aggressive truncation, retaining only tokens with higher confidence.\footnote{Similar to Eq.~\ref{eq:contrastive_prob}, we compute this constraint using logit value instead of post-softmax probability.}

By combining the contrastive probability distribution with the adaptive plausibility constraint, the $t$-th token is computed as
\begin{equation}
\begin{aligned}
p(y_t | \bm{x}, \bm{z}_d, \bm{z}_s, \bm{y}_{<t}) \quad \text{s.t.} \quad y_t \in \mathcal{V}_{\text{head}}(\bm{y}_{<t}).
\end{aligned}
\end{equation}

\section{Experiments on POPE}
\label{sec:experiment}


\paragraph{POPE}
We evaluate hallucination using the POPE benchmark~\cite{li2023evaluating}, which contains images from MSCOCO~\cite{MSCOCO}, A-OKVQA~\cite{A-OKVQA}, and GQA~\cite{GQA}. Models must answer yes/no questions probing for objects that are absent, generated via \textit{random}, \textit{popular} (high-frequency), and \textit{adversarial} (co-occurring) sampling. We report the average Accuracy, Precision, Recall, and F1 scores over five runs.
\paragraph{Models and Baseline}
To evaluate the robustness and adaptability of our methods, we select three state-of-the-art MLLMs with diverse architectures: LLaVA-v1.5-7B~\cite{liu2024improved}, Cambrian-8B~\cite{tong2024cambrian}, and MoLmo-7B-D~\cite{deitke2024molmo}. LLaVA-v1.5-7B utilizes a CLIP ViT-L/14@336~\cite{CLIP} vision encoder and the Vicuna-v1.5-7B~\cite{zheng2023judging} LLM. Cambrian-8B combines multiple vision encoders (e.g., CLIP ViT-L/14@336~\cite{CLIP}, ConvNeXt-XXL@1024~\cite{woo2023convnext}, DINOv2 Giant~\cite{oquab2023dinov2}, and SigLIP ViT-SO400M/14@384~\cite{zhai2023sigmoid}) with the LLaMA3-8B-Instruct~\cite{dubey2024llama} LLM. We chose MoLmo-7B-D, which pairs a CLIP ViT-L/336~\cite{CLIP} encoder with the QWen2-7B~\cite{yang2024qwen2} LLM, due to its superior performance over QWen-VL~\cite{bai2023qwen} on 11 benchmarks.



For baselines, we compare LayerCD with regular decoding and VCD~\cite{leng2024mitigating}. Please refer to Appendix for discussions on the differences between LayerCD and VCD.

\paragraph{Implementation Details}
For LayerCD, we set the contrastive amplification parameter $\alpha$ to 1 and the plausibility constraint's truncation parameter $\beta$ to 0.1. The shallow features $\bm{z}_s$ and deep features $\bm{z}_d$ are extracted from the first and last default layer of the vision encoder, respectively. The output tokens are sampled from the post-softmax layer after the decoding strategy is applied.



\begin{table*}[!t]
\tabstyle{7pt}
\begin{tabular}{llllll|l}
\toprule
 \textbf{Setting} & \textbf{Model} & \textbf{Decoding} & \textbf{Accuracy} & \textbf{Precision} & \textbf{Recall} & \textbf{F1 Score}  \\ \midrule

\multirow{9}{*}{\textit{Random}} & \multirow{3}{*}{LLaVA1.5} & Regular & $83.21_{\pm0.49}$ & $92.21_{\pm0.70}$ & $72.55_{\pm0.63}$ & $81.20_{\pm0.56}$ \\
  & & VCD & $82.33_{\pm0.16}$ & $95.42_{\pm0.52}$ & $67.93_{\pm0.30}$ & $79.36_{\pm0.17}$ \\
  & & LayerCD & $\textbf{85.77}_{\pm0.25}$ & $\textbf{96.39}_{\pm0.32}$ & $\textbf{74.32}_{\pm0.59}$ &  $\textbf{83.93}_{\pm0.34}$ \\[5pt]
  & \multirow{3}{*}{Cambrian} & Regular & $62.47_{\pm0.40}$ & $81.22_{\pm1.58}$ & $32.45_{\pm0.55}$ & $46.37_{\pm0.56}$
\\
  & & VCD & $64.59_{\pm0.56}$ & $94.54_{\pm1.43}$ & $30.96_{\pm0.82}$ & $46.64_{\pm1.04}$ \\
  & & LayerCD & $\textbf{75.65}_{\pm0.33}$ & $\textbf{97.21}_{\pm0.22}$ & $\textbf{52.81}_{\pm0.75}$ & $\textbf{68.44}_{\pm0.60}$ \\[5pt]
  & \multirow{3}{*}{Molmo} & Regular & $45.67_{\pm4.08}$ & $39.87_{\pm8.55}$ & $18.85_{\pm6.24}$ & $25.54_{\pm7.44}$ \\
  & & VCD & $50.29_{\pm0.47}$ & $50.58_{\pm0.96}$ & $24.80_{\pm1.04}$ & $33.28_{\pm1.14}$ \\
  & & LayerCD & $\textbf{60.79}_{\pm0.05}$ & $\textbf{63.49}_{\pm0.11}$ & $\textbf{50.81}_{\pm0.56}$ & $\textbf{56.45}_{\pm0.30}$ \\
  \midrule

  \multirow{9}{*}{\textit{Popular}} & \multirow{3}{*}{LLaVA1.5} & Regular & $81.83_{\pm0.47}$ & $88.99_{\pm0.53}$ & $72.64_{\pm0.66}$ & $79.99_{\pm0.55}$ \\
  & & VCD & $80.99_{\pm0.17}$ & $91.87_{\pm0.57}$ & $68.01_{\pm0.28}$ & $78.16_{\pm0.16}$ \\
  & & LayerCD & $\textbf{84.25}_{\pm0.25}$ & $\textbf{92.60}_{\pm0.17}$ & $\textbf{74.44}_{\pm0.59}$ & $\textbf{82.53}_{\pm0.34}$ \\[5pt]
  & \multirow{3}{*}{Cambrian} & Regular & $61.18_{\pm0.56}$ & $76.19_{\pm1.84}$ & $32.56_{\pm0.55}$ & $45.61_{\pm0.64}$ \\
  & & VCD & $62.91_{\pm0.68}$ & $88.19_{\pm1.28}$ & $29.81_{\pm1.18}$ & $44.55_{\pm1.42}$ \\
  & & LayerCD & $\textbf{73.87}_{\pm0.30}$ & $\textbf{91.45}_{\pm0.71}$ & $\textbf{52.68}_{\pm0.68}$ & $\textbf{66.84}_{\pm0.50}$ \\[5pt]
  & \multirow{3}{*}{Molmo} & Regular & $44.78_{\pm3.23}$ & $37.52_{\pm7.52}$ & $16.43_{\pm4.45}$ & $22.83_{\pm5.68}$ \\
  & & VCD & $49.21_{\pm0.32}$ & $48.22_{\pm0.77}$ & $21.47_{\pm0.77}$ & $29.70_{\pm0.89}$ \\
  & & LayerCD & $\textbf{58.83}_{\pm0.33}$ & $\textbf{61.60}_{\pm0.45}$ & $\textbf{46.92}_{\pm0.26}$ & $\textbf{53.27}_{\pm0.34}$ \\
  \midrule
                 
  \multirow{9}{*}{\textit{Adversarial}} & \multirow{3}{*}{LLaVA1.5} & Regular & $79.04_{\pm0.43}$ & $83.35_{\pm0.41}$ & $72.57_{\pm0.62}$ & $77.59_{\pm0.50}$ \\
  & & VCD & $78.32_{\pm0.35}$ & $85.72_{\pm0.93}$ & $67.97_{\pm0.33}$ & $75.82_{\pm0.26}$ \\
  & & LayerCD & $\textbf{82.09}_{\pm0.43}$ & $\textbf{87.96}_{\pm0.82}$ & $\textbf{74.36}_{\pm0.50}$ & $\textbf{80.59}_{\pm0.42}$ \\[5pt]
  & \multirow{3}{*}{Cambrian} & Regular & $61.23_{\pm0.44}$ & $76.66_{\pm1.45}$ & $32.31_{\pm0.67}$ & $45.45_{\pm0.69}$ \\
  & & VCD & $62.31_{\pm0.36}$ & $84.67_{\pm1.71}$ & $30.12_{\pm1.32}$ & $44.40_{\pm1.29}$ \\
  & & LayerCD & $\textbf{73.97}_{\pm0.28}$ & $\textbf{91.94}_{\pm0.56}$ & $\textbf{52.56}_{\pm0.66}$ & $\textbf{66.88}_{\pm0.50}$ \\[5pt]
  & \multirow{3}{*}{Molmo} & Regular &$44.81_{\pm3.29}$ & $39.05_{\pm6.78}$ & $19.25_{\pm4.51}$ & $25.77_{\pm5.50}$ \\
  & & VCD & $48.43_{\pm0.35}$ & $46.87_{\pm0.77}$ & $23.59_{\pm0.96}$ & $31.38_{\pm1.00}$ \\
  & & LayerCD & $\textbf{58.06}_{\pm0.62}$ & $\textbf{59.92}_{\pm0.76}$ & $\textbf{48.69}_{\pm0.62}$ & $\textbf{53.73}_{\pm0.68}$ \\
 \bottomrule
\end{tabular}
\caption{Results on POPE-MSCOCO. Our LayerCD outperforms VCD and regular decoding by significant margins. The results on POPE-A-OKVQA and POPE-GQA are provided in the supplementary where the conclusion remains the same.}
\label{tab:pope}
\end{table*}


\paragraph{Results on POPE}
Table~\ref{tab:pope} summarizes the results of applying different decoding strategies to different base models in the three distinct sampling settings. Overall, LayerCD achieves robust performance across all sampling settings and with different base models. Compared with regular decoding and VCD, LayerCD gains significant improvements consistently across all settings. These results strongly demonstrate the effectiveness of using shallow features to filter out object hallucinations (POPE only focuses on object hallucinations). It is worth noting that all models have relatively weak performance in terms of recall. This is probably due to the training data bias that causes the model to answer ``No''. Nonetheless, LayerCD still demonstrates huge gains against the baselines.

\section{Conclusion and Limitations}
\label{sec:Conclusion and Limitation}



In this work, we propose Layer Contrastive Decoding (LayerCD), a simple and effective approach built on the key observation that shallow visual features are significantly more prone to inducing MLLM hallucinations than deep ones. By leveraging the contrast between these feature levels, LayerCD effectively pinpoints and cancels out erroneous content.

A primary limitation of our approach is the doubled computational cost, as it requires separate forward passes for shallow and deep features—a drawback inherited from the original contrastive decoding design. This presents a challenge for extremely large models, and we leave the development of more efficient contrastive frameworks for future work.

\section{Acknowledgements}
This research is supported by Hong Kong Research Grants Council Early Career Scheme (No. 22200824).

{
    \small
    \bibliography{main}
    \bibliographystyle{plainnat}
}
\newpage
\appendix
\section{Appendix}
\subsection{Related Work}
\label{sec:related_work}

\paragraph{Hallucination in LLMs}
The causes of hallucinations in LLMs are multifaceted, with studies linking them to both training data quality~\cite{lin2021truthfulqa} and architectural limitations~\cite{li2023batgpt, liu2024exposing}. For instance, heuristically paired data during dataset construction can sometimes result in inconsistent or unsupported outputs, exacerbating the hallucination issue~\cite{lebret2016neural, wiseman2017challenges}. Additionally, limited diversity in training data, such as repetitive patterns~\cite{lee2021deduplicating}, can bias model outputs and increase the likelihood of hallucinations. Another significant cause for hallucination lies within the model architecture, where issues in representation learning and token embedding processing can distort models' understanding and amplify hallucinations~\cite{parikh2020totto}. Furthermore, models often prioritize memorized parametric knowledge over real-time input due to reliance on information encoded during training, further compounding the problem~\cite{roberts2020how, longpre2021entity}. Decoding strategies also play a role, with some methods introducing early-generation errors that accumulate rather than being corrected~\cite{zhang2023how, lee2022factuality, bengio2015scheduled}.

\paragraph{Hallucination in MLLMs}
The causes of hallucinations in MLLMs are more complex than in LLMs due to the integration of visual inputs. Specifically, limitations in the vision encoder, such as low resolution or a bias toward salient objects, have been identified as major contributors to hallucinations~\cite{zhai2023halle, li2024monkey, jain2024vcoder}. Additionally, the alignment process between visual and textual representations often fails to accurately synchronize these inputs, further exacerbating hallucination errors~\cite{jiang2024hallucination, chen2024internvl}. Furthermore, when visual input is incorporated into the model’s self-attention mechanism, it often receives insufficient focus, causing the model to rely more on pre-trained knowledge within the LLM component than on the actual visual content~\cite{favero2024multi, leng2024mitigating}.

Recent efforts to reduce hallucinations have employed various strategies, including the development of fine-grained datasets for improved training~\cite{gunjal2024detecting, liu2023mitigating}, enhancements to the vision encoder~\cite{jain2024vcoder}, better alignment mechanisms~\cite{jiang2024hallucination}, and optimized decoding strategies~\cite{huang2024opera, leng2024mitigating, wang2024mitigating}. Post-processing methods have also been utilized to address hallucinations after generation~\cite{zhou2023analyzing}, and reinforcement learning approaches have been explored to align models more closely with human preferences, improving the accuracy of generated outputs~\cite{zhao2023beyond, sun2023aligning}.

The most closely related work is Visual Contrastive Decoding (VCD)~\cite{leng2024mitigating}, which essentially contrasts output distributions generated from the original visual input with those generated from input distorted by Gaussian noise. Compared to VCD, our LayerCD approach addresses the problem from a novel perspective: we leverage shallow features to filter out hallucinations. In the experiments, we demonstrate that LayerCD significantly outperforms VCD on two challenging benchmarks. From the computation perspective, LayerCD performs favorably against VCD: VCD requires two forward passes in the vision encoder (one normal input and one distorted input) while LayerCD only needs one.


\subsection{Experiments on MME}
\paragraph{MME}
The MME benchmark~\cite{fu2024mmecomprehensiveevaluationbenchmark} provides a comprehensive toolbox to evaluate a wide range of capabilities in MLLMs, with a focus on perception and cognitive skills. This benchmark includes 14 different tasks, 10 of which assess perception-related abilities, while the remaining 4 are focused on cognitive processing. Following prior work~\cite{leng2024mitigating}, we select the \textit{existence} and \textit{count} subsets to assess object-level hallucinations and the \textit{position} and \textit{color} subsets to evaluate hallucinations related to object attributes. Similar to POPE, MME contains binary questions that require ``Yes'' and ``No'' responses. Model performance is quantified using a custom scoring formula, which aggregates various accuracy metrics to provide an overall assessment. To ensure fairness, the results reported are averaged over five runs.

\begin{table*}[!thbp]
\tabstyle{7pt}
\begin{tabular}{@{}llcccc@{}}
\toprule
 &  & \multicolumn{4}{c}{\textbf{MME subset}} \\
\textbf{Model} & \textbf{Decoding} & \textbf{Existence} & \textbf{Count} & \textbf{Position} & \textbf{Color} \\
\midrule
                              
\multirow{3}{*}{LLaVA1.5}   & Regular   & $169.67_{\pm3.71}$  & $113.67_{\pm7.18}$  & $117.67_{\pm10.73}$ & $140.67_{\pm6.55}$ \\
& VCD& $168.67_{\pm11.99}$ & $115.33_{\pm11.03}$ & $111.00_{\pm9.10}$  & $143.00_{\pm9.74}$ \\ 
& LayerCD & $\textbf{176.00}_{\pm3.74}$  & $\textbf{119.00}_{\pm13.97}$ & $\textbf{133.33}_{\pm11.01}$ & $\textbf{157.00}_{\pm5.10}$   \\
\midrule
                            
\multirow{3}{*}{Cambrian}   & Regular   & $98.67_{\pm8.06}$   & $85.67_{\pm18.15}$  & $66.33_{\pm7.48}$ & $74.33_{\pm8.34}$ \\
& VCD & $121.33_{\pm10.19}$ & $82.67_{\pm17.47}$  & $69.33_{\pm8.60}$ & $81.33_{\pm9.45}$ \\
 & LayerCD & $\textbf{125.33}_{\pm10.19}$ & $\textbf{88.67}_{\pm9.80}$ & $\textbf{75.68}_{\pm8.73}$ & $\textbf{103.67}_{\pm5.81}$   \\
\midrule
                            
\multirow{3}{*}{Molmo}  & Regular & $78.33_{\pm0.00}$   & $73.33_{\pm0.00}$   & $53.33_{\pm0.00}$ & $48.33_{\pm0.00}$ \\
& VCD & $70.33_{\pm3.40}$ & $\textbf{75.00}_{\pm2.58}$   & $53.67_{\pm4.40}$ & $\textbf{56.67}_{\pm2.36}$   \\
& LayerCD& $\textbf{78.33}_{\pm0.00}$ & $63.33_{\pm0.00}$& $\textbf{56.67}_{\pm0.00}$ & $53.33_{\pm0.00}$   \\
\bottomrule
\end{tabular}
\caption{Results on the MME hallucination subsets. LayerCD beats the two baselines, i.e., regular decoding and VCD, in most cases except the \textit{count} and \textit{color} subsets when Molmo is used as the base model.}
\label{tab:mme}
\end{table*}

\paragraph{Results on MME}
The MME hallucination subsets have been widely used by the hallucination research community. The results on these subsets are presented in Table~\ref{tab:mme} (left). Overall, LayerCD performs the best among the three decoding methods. In most settings, LayerCD outperforms the baselines with significant margins. In the \textit{existence} and \textit{position} tasks, LayerCD's gains are more significant. However, when using Molmo in the \textit{count} and \textit{color} tasks, LayerCD shows inferior results than VCD. In particular, we observe that the percentage of ``No'' answers in these two subsets is exceptionally high for Molmo and LayerCD somehow exacerbates this problem, resulting in weaker performance.

\subsection{Further Analysis}

\paragraph{Combining LayerCD with Traditional Decoding Strategies} 
Since LayerCD is theoretically orthogonal to the traditional decoding strategies, we combine them with LayerCD to see the effects. Specifically, we try the following decoding strategies: greedy decoding~\cite{germann2003greedy}, beam search~\cite{freitag2017beam}, and regular sampling with top-p~\cite{holtzman2019curious}, top-k, and temperature normalization. We conduct the experiments using LLaVA 1.5 on POPE based on COCO dataset and using random sampling setting. The results are summarized in Table~\ref{tab:decoding_stratiges}. It is clear that LayerCD works well with all the traditional decoding strategies, with top-p, top-k, and temperature normalization demonstrating the strongest synergy (the gains are significant).


\begin{table*}[h]
    \centering 
    \tabstyle{10pt}
    \begin{tabular}{lcc}
    \toprule
    Decoding Strategy & w/ LayerCD & Accuracy \\
    \midrule
    \multirow{2}{*}{Greedy} &  & $83.19$\\
     & \checkmark & $\textbf{85.67}$\\
     \midrule
    \multirow{2}{*}{\makecell[l]{Top P\\($P=0.9$)}} &  & $82.96$\\
     & \checkmark & $\textbf{84.88}$\\
     \midrule
    \multirow{2}{*}{\makecell[l]{Top K\\($K=50$)}} &  & $83.02$\\
     & \checkmark & $\textbf{85.23}$\\
     \midrule
    \multirow{2}{*}{\makecell[l]{$\text{Top K \& Temperature}$ \\ ($K=50; \text{temperature}=0.7$)}} &  & $84.04$\\
     & \checkmark & $\textbf{85.36}$\\
     \midrule
    \multirow{2}{*}{\makecell[l]{$\text{Top K \& Temperature}$ \\ ($K=50; \text{\textit{temperature}}=1.5$)}} &  & $83.95$\\
     & \checkmark & $\textbf{85.02}$\\
    \midrule
    \multirow{2}{*}{\makecell[l]{$\text{Beam Search}$ \\ ($Num_{beams}=3$)}} &  & $84.20$\\
     & \checkmark & $\textbf{86.11}$\\
    \bottomrule
    \end{tabular}
    \caption{Results of combining different decoding strategies with LayerCD. LayerCD works well with all of them.}
    \label{tab:decoding_stratiges}
\end{table*}

\paragraph{Impact of Hyperparameters $\alpha$ and $\beta$}

Table~\ref{tab:ablation_hyper_parameter} shows the results of varying $\alpha$ and $\beta$, which are LayerCD's hyperparameters for controlling the contrastive amplification and constraint truncation, respectively. As $\alpha$ increases from $0.2$ to $1.0$, the performance of LayerCD improves accordingly. The results suggest that a higher $\alpha$ facilitates the amplification of the difference between the two output distributions derived from shallow and deep visual features, respectively, and hence leads to better performance. For $\beta$, the results do not have a clear pattern. Though a higher $\beta$ leads to better performance, we find in practice that setting $\beta$ too high may reduce output diversity.

\begin{table}[!t]
\begin{center}
\tabstyle{7pt}
\begin{tabular}{@{}c@{\hspace{1cm}}c@{}}
\begin{tabular}[b]{c|c}
\toprule
$\alpha$ & Accuracy \\
\midrule
$0.2$ & $83.7$\\
$0.4$ & $84.18$\\
$0.6$ & $85.33$\\
$0.8$ & $82.45$\\
$1.0$ & $\textbf{85.67}$\\
\bottomrule
\end{tabular}
&
\begin{tabular}[b]{c|c}
\toprule
$\beta$ & Accuracy \\
\midrule
$0.001$ & $81.96$\\
$0.01$ & $83.01$\\
$0.1$ & $\textbf{85.67}$\\
$0.2$ & $84.75$\\
$0.5$ & $81.96$\\
$0.9$ & $80.58$\\
\bottomrule
\end{tabular}
\\
(a) & (b)
\end{tabular}
\caption{Impact of hyperparameters $\alpha$ and $\beta$. See Methodology section for their uses.}
\label{tab:ablation_hyper_parameter}
\end{center}
\end{table}


\begin{wraptable}{r}{0.3\textwidth}
\tabstyle{5pt}

\vspace{-10pt} 
\begin{tabular}{ccccc}
\toprule
w/ APC & \textbf{Average} \\
\midrule
  & $62.21$\\
\checkmark &  $85.67$\\ 
\bottomrule
\end{tabular}
\caption{Impact of the Adaptive Plausibility Constraint. \textit{APC} denotes the Adaptive Plausibility Constraint.}

\vspace{-15pt} 
\label{tab:apc}
\end{wraptable} 
\paragraph{Impact of Adaptive Plausibility Constraint}
Table~\ref{tab:apc} presents the LayerCD results of LLaVA1.5-7B on POPE based on COCO dataset and using random sampling setting, comparing performance with and without the Adaptive Plausibility Constraint. The results indicate that the Adaptive Plausibility Constraint plays a crucial role in enhancing LayerCD's performance.

A higher $\alpha$ amplifies the difference between the two output distributions derived from shallow and deep visual features and makes desirable tokens easier to be selected. Notably, because POPE responses are often single tokens and the highest-probability token is usually correct, a higher $\beta$ can filters out more distractor tokens and increases performance. However, setting $\beta$ too high may reduce output diversity.

\end{document}